\documentclass{article}

    \PassOptionsToPackage{numbers, compress}{natbib}

\usepackage[final]{neurips_glfrontiers_2022}

\usepackage[utf8]{inputenc} 
\usepackage[T1]{fontenc}    
\usepackage{hyperref}       
\usepackage{url}            
\usepackage{booktabs}       
\usepackage{amsfonts}       
\usepackage{nicefrac}       
\usepackage{microtype}      
\usepackage{xcolor}         

\usepackage{amsmath}

\usepackage{tabularx}
\usepackage{multirow}
\usepackage{graphicx}

\usepackage{algorithm}
\usepackage{algorithmic}
\makeatletter
\newcommand{\ALOOP}[1]{\ALC@it\algorithmicloop\ #1%
  \begin{ALC@loop}}
\newcommand{\ENDALOOP}{\end{ALC@loop}\ALC@it\algorithmicendloop}

\makeatother

\title{Improving Graph Neural Networks  at Scale: Combining Approximate PageRank and CoreRank
}

\author{%
  Ariel R. Ramos Vela\thanks{Corresponding authors: \texttt{ariel.ram97@gmail.com; johannes.lutzeyer@polytechnique.edu}.} \\ Télécom Paris, IP Paris  \\ Palaiseau, France \And Johannes F. Lutzeyer$^*$ \\ LIX, École Polytechnique, IP Paris \\ Palaiseau, France \And  Anastasios Giovanidis\\ Sorbonne University CNRS-LIP6 \\ Paris, France 
  \And Michalis Vazirgiannis \\ LIX, École Polytechnique, IP Paris \\ Palaiseau, France
}

\begin{document}

\maketitle

\begin{abstract}
Graph Neural Networks (GNNs) have achieved great successes in many learning tasks performed on graph structures. Nonetheless, to propagate information GNNs rely on a message passing scheme which can become prohibitively expensive when working with industrial-scale graphs. Inspired by the PPRGo model, we propose the CorePPR model, a scalable solution that utilises a learnable convex combination of the approximate personalised PageRank and the CoreRank to diffuse multi-hop neighbourhood information in GNNs. Additionally, we incorporate a dynamic mechanism to select the most influential neighbours for a particular node which reduces training time while preserving the performance of the model. Overall, we demonstrate that CorePPR outperforms PPRGo, particularly on large graphs where selecting the most influential nodes is particularly relevant for scalability. Our code is publicly available at: \url{https://github.com/arielramos97/CorePPR}.
\end{abstract}

\section{Introduction}\label{sec:intro}

The rapid development of Graph Neural Networks (GNNs) for solving tasks such as semi-supervised classification, regression and link prediction has led to the development of high-performance models that exploit message passing, i.e., neighbourhood aggregation, mechanisms to extract and propagate information in a graph. However, at the extreme scale that current businesses are faced with, this mechanism is often too expensive. 
Previous work \cite{Gasteiger2019}, demonstrates that the personalised PageRank (PPR) can be employed to compute the influence scores of nodes with respect to a source node and thus, directly include the information of multi-hop neighbourhoods within a single message passing step. However, this approach becomes infeasible when working with large graphs since it uses a variant of Power Iteration to approximate personalised PageRank scores when training. This is highly costly even for a few iterations given that they are computed during each gradient-update step.
Hence, \cite{Bojchevski2020} proposed the PPRGo model, which computes a sparse approximation of the personalised PageRank vectors based on the Push algorithm by \cite{Andersen2006}. The PPRGo model is able to trade-off scalability and performance while being able to train and test on their newly introduced MAG-Scholar dataset (10.5M nodes, 133M edges, 2.8M node features) in under 2 minutes.

\textbf{Contributions.} Motivated by this model, the main contribution of our work is to introduce CorePPR, a GNN model that makes use of a learnable convex combination of both CoreRank \cite{Tixier2016} and the approximated personalised PageRank. Here, we effectively learn a more efficient graph structure over which to propagate information. In this learned graph structure, links are drawn based on approximated personalised PageRank values and then weighted based on a learnable linear combination of the personalised PageRank and the CoreRank. Hence, the learned graph structure in our approach differentiates between the selection of neighbours and the weighting of the connections to those neighbours. In the selection phase, we only use the approximate personalised PageRank (similar to PPRGo) to consider a local perspective of each node with respect to the rest of the graph; whereas in the weighting phase, we use the convex combination including CoreRank, which is a global score of each node's importance, in order to combine the local with the global perspective using a learnable weighting between the two. 
Since, the calculation of CoreRank scores is linear in the sum of the number of nodes and edges, the scalability of the PPRGo model, which solely relied on the highly scalable Push method to approximate personalised PageRank scores, is maintained~in~our~CorePPR~model.

The PPRGo model \cite{Bojchevski2020} treats the number of neighbours to consider in the propagation process as a hyperparameter with fixed value $\ell$ (in \cite{Bojchevski2020} the notation was $k$) for all nodes. In our work, we include a dynamic mechanism selecting the number of neighbours for any given node $i,$ denoted by $\ell_i,$ which can be different for each node. 
This allows us to avoid manually tuning the hyperparameter $\ell$ and reduces the computational cost of the model, while maintaining its performance. Importantly, our two contributions, the learnable weight between local and global perspective and the adaptive number of neighbours, do not rely on each other and, especially the second of these, could speed up computation in several existing GNN models, such as the Graph Attention Networks \cite{Velivckovic2018,Brody2021}.

\section{Background and Related Work} \label{sec:bgrw}


First, we present the notation used in our model. Let $n$ denote the number of nodes and $m$ the number of edges of a graph $G.$ The feature matrix is described by $X\in \mathbb{R}^{n\times f},$ with the number of features $f,$ while the one-hot class matrix is described by $Y \in \mathbb{R}^{n\times c},$ with the number of classes $c.$ We furthermore denote the $n\times n$ identity matrix by $I_n.$ Finally, the adjacency matrix is denoted by $A,$ while $\tilde{A} = (D+I_n)^{-1/2} (A+I_n) (D+I_n)^{-1/2}$ denotes the normalised adjacency matrix with  added self-loops and $D$ is the corresponding degree matrix. 
We now review different GNNs models that utilise PageRank scores in their message passing scheme. 



\textbf{From Message Passing to Personalised PageRank.} Due to the scalability limitation of message passing architectures, \cite{Gasteiger2019} proposed the PPNP model separating the neural network used to generate~predictions~from~the~propagation~scheme. It is defined as follows,
\begin{equation}\label{eqn:ppnp}
Z= \mathrm{softmax}\left(\Pi_{sym}H\right), \qquad H=f_\theta(X), \qquad \Pi_{sym}=\alpha (I_n-(1-\alpha)\tilde{A})^{-1},
\end{equation}
where each row in the propagation matrix $\Pi_{sym}$ 
corresponds to the personalised PageRank scores of a given node,  $\alpha \in (0, 1]$ is the teleport (or restart) probability in the personalised PageRank algorithm, $H$ is the node representation matrix generated by $f_\theta$, which  refers to a neural network with parameters $\theta$, whereas $Z \in \mathbb{R}^{n\times c}$ is the prediction matrix.
This model effectively incorporates information of multi-hop neighbourhoods in one single step. 
However, computing the dense matrix $\Pi_{sym}$ for all nodes is expensive in terms of time and memory. Hence, Power Iteration is used during inference of the PNPP model. But even a few iterations on large graphs are still computationally costly. 


\textbf{From Personalised PageRank to Approximate PageRank.} A fast method that can efficiently compute PageRank for any set of nodes having a score above some predefined threshold is the Push method, introduced in \cite{Andersen2006}. Crucially, this method identifies a subset of nodes for which to approximate PageRank scores with a tuneable precision and avoids the computation of scores for all nodes in a potentially large graph. 
Therefore, instead of computing $\Pi_{sym},$ in the PPRGo model \cite{Bojchevski2020} proposed to calculate a sparse approximation $\Pi_\epsilon(\ell\mathbf{1})$ of $\Pi_{ppr} = \alpha (I_n-(1-\alpha)D^{-1}A)^{-1}$ by employing the Push method which offers scalability but trades-off precision of the scores \cite{Wu2021}.
\begin{equation}\label{eqn:appnp}
Z= \mathrm{softmax}\left(\Pi_{\epsilon}(\ell\mathbf{1})H\right), \qquad H=f_\theta(X),
\end{equation}
where $\Pi_\epsilon(\ell\mathbf{1})$ is obtained via the Push algorithm, run with precision $\epsilon,$ \cite{Andersen2006} and then setting all but the $\ell$ largest entries in every row to zero. 
To be precise, the non-zero entries in a given row $i$ of $\Pi_\epsilon(\ell\mathbf{1})$ are the $\ell$ largest approximated personalised PageRank scores teleporting to, i.e., restarting from, reference node $i.$
This model allows to trade-off scalability and performance by varying the number of chosen neighbours $\ell$ and utilising the feature representations $f_\theta(x_i)$ of only these top-$\ell$ neighbours during inference. Nonetheless, $\ell$ is a hyperparameter that is fixed meaning that it needs to be tuned.

\section{Methodology} \label{sec:methodology}
Our CorePPR model is inspired by the scalability and performance trade-off of PPRGo and incorporates the CoreRank \cite{Tixier2016}, a centrality metric used to score nodes based on the $k$-core decomposition of a graph. We decide to use CoreRank for two main reasons. First, the CoreRank can be efficiently calculated in $\mathcal{O}(2m)$ and has no precision issues since its values are discrete unlike PageRank. 
Secondly, via CoreRank we include a global perspective of the importance of nodes and thus, complement the local perspective of the approximate personalised PageRank. 

The \textit{$k$-core} of a graph \cite{Malliaros2020, Tixier2016} is defined to be the maximal connected subgraph of a graph $G$ in which all vertices have at least degree $k.$ Then, the \textit{core number} of a node is computed by selecting the largest value of $k$ such that there exists a $k$-core subgraph containing the given node.  
To obtain \textit{CoreRank scores} \cite{Tixier2016} we sum core numbers over neighbourhoods (see Appendix \ref{app:algos} Algorithm \ref{algo:CoreRank}).  


Similar to PPRGo, our model decouples the feature transformation from the information propagation. 
The main innovation in our CorePPR model lies in the definition of the message passing operator $\Pi_\gamma,$
\begin{equation}\label{eqn:CorePPR}
Z= \mathrm{softmax}\left(\Pi_{\gamma}H\right), \qquad H=f_\theta(X), \qquad \Pi_\gamma = (1-\gamma)  \Pi_\epsilon({\boldsymbol \ell}) + \gamma  \mathrm{C}.
\end{equation}
The learnable parameter $\gamma \in [0, 1)$, weights the information of a local perspective of the graph in $\Pi_\gamma({\boldsymbol\ell})$ and from a global perspective of the graph in $C.$ 
For values of $\gamma$ close to 0 we give more importance to the local (approximate personalised PageRank) scores whereas for values close to 1, the global (coreRank) metric predominates. To find an optimal value of $\gamma$, we defined it as a trainable parameter and train it jointly with the neural network parameters $\theta$ via the backpropagation algorithm. 

Now we turn to the definitions of the matrices $\Pi_\epsilon({\boldsymbol \ell})$ and $ \mathrm{C}.$ Like $\Pi_\epsilon(\ell\mathbf{1}),$ our matrix $\Pi_\epsilon({\boldsymbol \ell})$ contains the approximated personalised PageRank scores obtained via the Push algorithm run with precision $\epsilon.$ The two matrices only differ in their sparsity pattern. While the non-zero entries in each row of $\Pi_\epsilon(\ell\mathbf{1})$ equal the largest $\ell$ approximated personalised PageRank scores, the  non-zero entries in each row of $\Pi_\epsilon({\boldsymbol \ell})$ equal the largest $\ell_i$ approximated personalised PageRank scores, i.e., we adaptively choose the sparsification ratio of each node.
Given that some nodes are influenced by only a few neighbours while others are influenced by a large number of neighbours, taking a fixed $\ell$ may not provide enough information for some nodes, while it may add redundant information to others. Besides, empirically testing values for $\ell$ is costly. Hence, another contribution made by our algorithm is that instead of considering the top $\ell$ nodes, we use Appendix \ref{app:algos} Algorithm \ref{algo:elbow} to select the nodes before the elbow point in the curve formed by the approximate personalised PageRank scores.  
Note that this selection is purely based on personalised PageRank scores. The CoreRank metric is a global metric and in no way personalised and therefore, is unsuitable to introduce a sparsity pattern to the matrix  $\Pi_\epsilon({\boldsymbol \ell}),$  which  helps disambiguate nodes in the node classification task. The $(ij)^{\mathrm{th}}$ entry of the matrix $C$ equals to the normalised coreRank of node $j$ if the personalised PageRank score of node $j$ is among the ${\boldsymbol\ell}_i$ largest scores in the personalised PageRank restarting at node $i.$ Hence, the sparsity pattern of $\Pi_\gamma({\boldsymbol\ell})$ and $C$ is identical. Note that the rows of $C$ are normalised to sum to 1.

Like the PPRGo model, our CorePRR model can be run by precomputing both the approximated personalised PageRank and the CoreRank scores before training the model. 
Note that only 
$n$ CoreRank values, one per node, are required, while for the personalised PageRank several scores are computed per node, with the exact number being determined by the Push method.

\section{Experiments} \label{sec:experiments}

To demonstrate the effectiveness of CorePPR with respect to PPRGo, we performed experiments using the same hyperparameters to observe the performance of both models under the same conditions. Summary statistics of the utilised datasets can be found in Appendix \ref{app:experiments}. Our code is publicly available online \footnote[1]{\url{https://github.com/arielramos97/CorePPR}}. 
We answer the following research questions: 
\textbf{(RQ1)}~How effective is the incorporation of the global coreRank metric in our model?
\textbf{(RQ2)}~Is it possible to trade-off performance and scalability by using the proposed elbow algorithm?
\textbf{(RQ3)}~Are there any significant overheads induced by our model with respect to PPRGo?


\textbf{CorePPR variants. } We propose two versions of our CorePPR model. OT (Only Training) only utilises the convex combination of CoreRank and the personalised PageRank for training, the inference step is identical to the PPRGo model and does not make use of the CoreRank metric, but makes use of neural network parameters which were trained in the presence of CoreRank scores. For reasons of computational efficiency, the inference in the  PPRGo model is performed by estimating class probabilities using Power Iteration to implicitly make use of personalised PageRank approximations without ever obtaining these PageRank scores. 
Alternatively, in the T\&T (Training and Testing) variant of our PPRGo model we utilise the convex combination of both metrics during training and inference. In terms of computational cost, T\&T is more expensive given that it needs to explicitly compute the approximate personalised PageRank scores for all nodes during inference.

\textbf{Complexity.} 
In the following we denote the number of nodes in the training  and test set by $n_{train}$ and $n_{test},$ respectively, and the number of edges in the test set by $m_{test}.$
As for the PPRGo model the complexity of a \textit{forward pass} through our model (with fixed $\ell$) during training is $\mathcal{O}(n_{train} (f+ \ell) c),$ since the addition of the CoreRank matrix only incurs the complexity of matrix addition, which is subsumed in the existing terms. 
The \textit{test complexity} of both the PPRGo model and the OT variant of our model is $\mathcal{O}(2m_{test}),$ since it requires two sparse matrix multiplications by the adjacency matrix to perform Power Iteration. In the T\&T variant of our model we instead need to run the Push algorithm and calculate the CoreRank of all nodes in the test set, which has a time complexity of $\mathcal{O}(\frac{1}{\epsilon \alpha}  n_{test} + 2 m),$ see \cite[Thm.~1]{Andersen2006} and  \cite{ Tixier2016}.

\textbf{Initialisation of $\gamma.$} We always initialise $\gamma$ at 0 such that it assigns 0.5, i.e., equal importance, to both approximate PageRank and CoreRank after being passed through the sigmoid function. 
In Table \ref{tab:results}, it can be observed that \textit{for larger datasets} the value of $\gamma$ is higher meaning that \textit{the global structure of a graph becomes more relevant}. E.g., for our largest dataset MAG-Scholar-C we observe a weight of $\gamma=0.45$ of the global CoreRank scores, which is significantly higher than the values of $\gamma$ observed for the other, smaller datasets. Interestingly, we see that for all datasets  there seems to be only one optimal value of $\gamma$ per dataset, clearly demonstrating that there exists an optimal balance between the local and global perspective for each dataset, which is reliably identified by our CorePPR model.

\begin{table*}[t]
\caption{Runtime(s), memory(GB), and accuracy for different models run with a Push algorithm precision $\epsilon=1e^{-4}$ and restart probability $\alpha=0.25$ (except for the Reddit dataset, for which $\alpha=0.5$ in line with the PPRGo experiments). For our CorePPR model, the optimal $\gamma$ as well as the mean number of neighbours in the column ``Mean $\ell$'' is displayed. 
The appendage to the model name ``\textbf{OT}'' denotes that we use the convex combination of CoreRank and approximate personalised PageRank Only during Training; whereas ``\textbf{T\&T}'' denotes that the convex combination is used in both Training and Testing. Finally, ``\textbf{d${\boldsymbol \ell}$}'' denotes that the model use the Dynamic mechanism for the selection of ${\boldsymbol \ell}.$ We set the best values in bold and report the average results and standard deviations of 25 repetitions of each experiment.    
} \label{tab:results}
\begin{center}\resizebox{\linewidth}{!}{
\begin{tabular}{rcccccccccc}
\hline
\hline
& \multicolumn{5}{c}{\textbf{Cora-Full}} &  \multicolumn{5}{c}{\textbf{PubMed}}    \\
&Mean $\ell$& $\gamma$ &Time & Memory & Accuracy & Mean $\ell$ & $\gamma$ & Time & Memory & Accuracy\\
\hline
\textbf{PPRGo  } & 32 & -- & 10.9(2.1) & \textbf{1.5(0.0)} & 61.35(0.99) & 32 & -- & \textbf{1.0(0.0)} & 5.3(0.0)       & 74.57(3.61) \\
\textbf{CorePPR  T\&T } & 32 & 0.09(0.00) & 13.3(2.7) & 4.6(0.1) & \textbf{61.83(0.94)} & 32 & 0.29(0.00) & 10.0(2.3) & 5.4(0.0) & \textbf{74.65(3.51)} \\
\textbf{CorePPR  OT } & 32 & 0.09(0.00) & \textbf{1.4(0.0)} & 11.3(2.1) & 61.60(0.97) & 32 & 0.29(0.01) & 1.4(0.0)  & 1.6(2.0)   & 74.43(3.63) \\
\textbf{CorePPR  T\&T d${\boldsymbol \ell}$ } & 16 & 0.10(0.00) & 4.3(0.1) & 8.1(3.0)  & 61.56(1.04) & 12 & 0.30(0.01) & 5.4(0.0) & 9.7(3.1) & 72.50(3.38) \\
\textbf{CorePPR  OT d${\boldsymbol \ell}$ } & 16 & 0.10(0.00)  & \textbf{1.4(0.0) }         & 7.6(3.3)  & 61.37(0.93) & 12              & 0.30(0.00)   & 1.4(0.0)          & \textbf{2.0(3.8)}    & 74.15(3.21) \\
\hline
&\multicolumn{5}{c}{\textbf{Reddit}}  & \multicolumn{5}{c}{\textbf{MAG-Scholar-C}}  \\
&Mean $\ell$& $\gamma$ & Time & Memory & Accuracy & Mean $\ell$ & $\gamma$ & Time & Memory & Accuracy\\
\hline
\textbf{PPRGo  } & 32 & -- & 9.3(0.5) & \textbf{5.3(0.0)} & 26.41(1.30) & 32 & -- & 262.2(4.0) & \textbf{28.0(0.5)} & 68.25(2.48) \\
\textbf{CorePPR  T\&T }& 32 & 0.17(0.04) & 5.3(12.2) & 6.1(0.3) & 33.19(0.81) &-- &-- &-- &-- &-- \\
\textbf{CorePPR  OT } & 32 & 0.17(0.04) & 5.6(0.0) & 13.0(1.8) & 29.61(1.00) & 32  & 0.44(0.01) & 251.3(3.8) & 28.3(0.6) &  \textbf{72.15(1.73)} \\
\textbf{CorePPR  T\&T d${\boldsymbol \ell}$} & 3  & 0.18(0.04) & 6.1(0.3) & 6.1(17.6) & \textbf{33.23(0.89)} &-- &-- &-- &-- &-- \\
\textbf{CorePPR  OT d${\boldsymbol \ell}$ } & 3 & 0.18(0.04)   & \textbf{5.2(0.1)}          & 12.6(4.3)  & 29.20(1.17) & 23              & 0.45(0.00)   & \textbf{245.2(3.1)}  & 28.8(0.6) & 71.86(1.59) \\
\hline
\end{tabular}}\\
\end{center}
\end{table*}



\textbf{Analysis of Results.} 
For the OT version of our model, the results in Table \ref{tab:results} are comparable to PPRGo when the datasets are small. Nonetheless, for larger datasets such as Reddit and MAG-Scholar-C, CorePPR outperforms PPRGo by $3\%$ (6990 nodes) and $4\%$ (420000 nodes), respectively. This difference is even more significant for the MAG-Scholar-C dataset since $1\%$ of the nodes is equivalent to approximately $105000$ nodes which is larger than the Cora-Full and PubMed datasets.
Similarly, for the T\&T version our CorePPR model performs slightly better than OT for small datasets, whereas for large datasets it \textit{outperforms PPRGo by a 7\% difference} on Reddit (approximately $16310$ nodes). Given that T\&T is computationally more expensive than OT, it was infeasible to compute the results for the MAG-Scholar-C dataset.
Overall, \textit{the inclusion of the global CoreRank metric in our CorePPR model impacts the performance achieved by PPRGo positively} in both versions OT and T\&T \textbf{(RQ1)}.

In Table \ref{tab:results} we observe that, when the number of neighbours is chosen dynamically,  both versions of the CorePPR model utilise significantly fewer nodes (2 and 3 on average on the Reddit dataset), while achieving similar or better performance than the PPRGo model, considering 32 neighbours \textbf{(RQ2)}. Utilising fewer nodes allows our model to train faster. This is more noticeable for larger datasets such as Reddit where both OT and T\&T versions (using the dynamic $\ell$ mechanism) take less time to run and achieve an even higher performance with respect to the PPRGo baseline. Likewise, for the MAG-Scholar-C dataset, the OT version of our model with dynamic $\ell$ is capable of achieving an accuracy of $72\%$  while having a shorter computing time. In terms of memory and time requirements, we demonstrate that the OT version of our CorePPR model does not incur significant overheads. On the other hand, the T\&T version is significantly more expensive than OT and becomes infeasible to apply for extremely large datasets such as MAG-Scholar-C \textbf{(RQ3)}.





\section{Conclusion}

CorePPR is a GNN model used for semi-supervised node classification that scales to large datasets. The results obtained by our model demonstrate that we can further improve the accuracy, by up to $7\%,$ achieved by PPRGo by including the coreRank global metric and using it in a  convex combination with the approximate personalised PageRank. Besides, this combination does not need to be manually tuned since gamma is a learnable parameter that changes depending on the dataset. 
In addition, we provide an adaptive mechanism that allows us to use the most significant $\ell_i$ neighbours of a particular node $i$ instead of using a fixed number of neighbours for all nodes as was done in the original PPRGo model. This in turn decreases the training time needed.

\section*{Acknowledgements}
The research of Ariel R. Ramos Vela and Anastasios Giovanidis is supported by the French National Agency of Research (ANR) through the FairEngine project under Grant
ANR-19-CE25-0011. The work of Dr. Johannes Lutzeyer and Prof. Michalis Vazirgiannis is supported by the ANR chair AML-HELAS (ANR-CHIA-0020-01).

\bibliographystyle{plain}
\bibliography{references}


\newpage
\appendix

\section*{Appendix}


\section{Algorithms Utilised in CorePPR} \label{app:algos}

In this section we outline both the CoreRank and Elbow Detection algorithms used in our CorePPR model. Here we use $\mathcal{N}(i)$ to denote the neighbourhood of node $i$ in a given graph $G.$ Note that in the Elbow algorithm we consider only the PageRank scores of a given node's neighbours and not its own PageRank score. The personalised PageRank score of the central node is often observed to be significantly larger than that of its neighbours. Hence, we empirically observed the results of our Elbow algorithm to improve when the personalised PageRank of the central node was not considered.

\begin{algorithm}[h]
   \caption{CoreRank}
   \label{algo:CoreRank}
\begin{algorithmic}[1]
   \STATE {\bfseries Input:} Core decomposition of $G$
   \STATE {\bfseries Output:} $C$ vector containing coreRank scores

    \STATE $C$ $\gets$ empty vector of length $n$ 
    
   \FOR{$i\gets1$ {\bfseries to} $n$}
    \STATE  $C[i]\gets \sum_{u \in \mathcal{N}(i)} \mathrm{corenumber}(u)$ 
    \ENDFOR
    \RETURN $C$
\end{algorithmic}
\end{algorithm}

\begin{algorithm}[h]
   \caption{Elbow \cite{Tixier2016}}
   \label{algo:elbow}
\begin{algorithmic}[1]
   \STATE {\bfseries Input:} Set of 2 dimensional points, $S = \{ (x_1, y_1), \ldots, (x_{| S|}, y_{| S|})\},$ of size $| S| \geq 2$
   \STATE {\bfseries Output:} $x_{elbow}$

    \STATE $line\gets \{ (x_0, y_0); (x_{| S|}, y_{| S|})\}$

   \IF{$| S| > 2$} 
        \STATE $distance \gets$ empty vector of length $|{S}|$ 
        \STATE $s\gets 1$

        \FOR{$(x,y)$ {\bfseries in} $S$} 
            \STATE $distance[s]\gets$ distance from $(x,y)$ to $line$
            \STATE $s\gets s+1$
        \ENDFOR
        \IF{$\exists !s | distance[s] = max(distance)$} 
            \STATE $x_{elbow}\gets x | (x,y)$ is most distant from $line$
        \ELSE
            \STATE $x_{elbow}\gets x_0$
        \ENDIF
    \ELSE
        \STATE $x_{elbow}\gets x | y$ is maximum
    \ENDIF
    \RETURN $x_{elbow}$

\end{algorithmic}
\end{algorithm}

\begin{figure*}[t]
    \centering
    \includegraphics[width=0.7\textwidth]{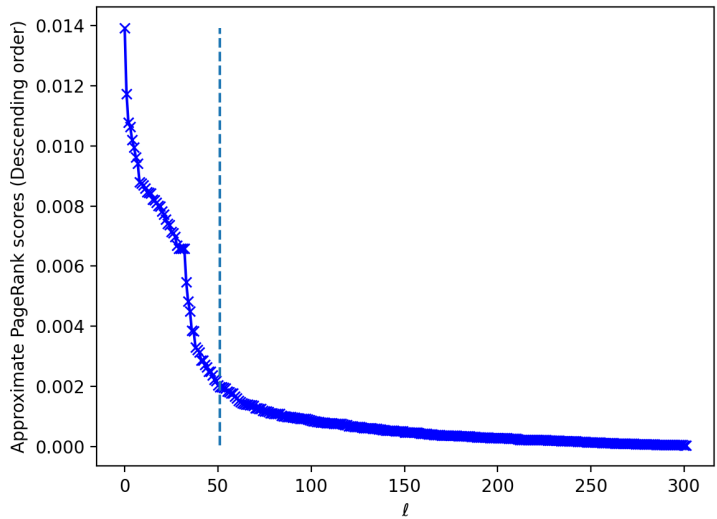}
    \caption{Illustration of the approximated personalised PageRank scores for a node in the Cora-Full dataset. The dashed vertical line illustrates the elbow point selected using Algorithm \ref{algo:elbow}.}
    \label{fig:elbow_examples}
\end{figure*}

\section{Datasets and Further Experiment Details}\label{app:experiments}

In this section we provide summary statistics of the datasets used in our experiments and a further set of experiments, where the Push algorithm was run with precision $\epsilon =1e^{-2}$ in Table \ref{tab:results_e2}. The results in Table \ref{tab:results_e2} support the conclusions drawn in the main paper. 

\textbf{Datasets.} To facilitate the comparison of our model with PPRGo, we utilised the same datasets: Cora-Full \cite{bojchevski2018deep} (18.7K nodes, 63.4K edges, 8.7K node features), PubMed \cite{DBLP:journals/kais/YangL15}(19.7K nodes, 44.3K edges, 0.5K node features), Reddit \cite{NIPS2017_5dd9db5e} (233K nodes, 11.6M edges, 602 node features) and the MAG-Scholar-C dataset \cite{Bojchevski2020} (10.5M nodes, 133M edges, 2.8M node features).

\begin{table*}[t]
\caption{Runtime(s), memory(GB), and accuracy for different models run with a Push algorithm precision $\epsilon=1e^{-2}$ and restart probability $\alpha=0.25$ (except for the Reddit dataset, for which $\alpha=0.5$ in line with the PPRGo experiments). For our CorePPR model, the optimal $\gamma$ as well as the mean number of neighbours in the column ``Mean $\ell$'' is displayed. 
The appendage to the model name ``\textbf{OT}'' denotes that we use the convex combination of CoreRank and approximate personalised PageRank Only during Training; whereas ``\textbf{T\&T}'' denotes that the convex combination is used in both Training and Testing. Finally, ``\textbf{d${\boldsymbol \ell}$}'' denotes that the model use the Dynamic mechanism for the selection of ${\boldsymbol \ell}.$ We set the best values in bold and report the average results and standard deviations of 25 repetitions of each experiment.
} \label{tab:results_e2}
\begin{center}\resizebox{\linewidth}{!}{
\begin{tabular}{rcccccccccc}
\hline
\hline
& \multicolumn{5}{c}{\textbf{Cora-Full}} &  \multicolumn{5}{c}{\textbf{PubMed}}    \\
&Mean $\ell$& $\gamma$ &Time & Memory & Accuracy & Mean $\ell$ & $\gamma$ & Time & Memory & Accuracy\\
\hline
\textbf{PPRGo  } & 32 & -- & 2.8(0.1)  & \textbf{1.5(0.0)}  & 58.61(1.17) & 32 & -- & \textbf{0.7(0.0)} & 5.3(0.0) & \textbf{73.42(4.00)} \\
\textbf{CorePPR  T\&T } & 32 & 0.11(0.00) & 3.7(2.3)  & \textbf{1.5(0.0)} & 53.60(0.93) & 32 & 0.30(0.01) & 1.7(2.1) & 1.5(0.0) & 70.78(3.44) \\
\textbf{CorePPR  OT } & 32 & 0.11(0.00) & \textbf{1.4(0.0)} & 3.8(2.2)  & \textbf{59.17(1.13)} & 32 & 0.30(0.01) & 1.4(0.0)  & \textbf{1.4(1.9)}  & 73.28(3.94) \\
\textbf{CorePPR  T\&T d${\boldsymbol \ell}$ } & 3  & 0.11(0.00) & 1.5(0.0) & 3.0(2.8)  & 51.94(1.15) & 3  & 0.30(0.01) & 1.6(0.0) & 1.7(2.8) & 69.43(3.98) \\
\textbf{CorePPR  OT d${\boldsymbol \ell}$ } & 3 & 0.11(0.00)   & \textbf{1.4(0.0)}          & 3.8(3.5)  & 57.17(1.21)                         & 3               & 0.30(0.01)   & 1.4(0.0)          & 1.9(3.6)    & 71.86(4.74) \\
\hline
&\multicolumn{5}{c}{\textbf{Reddit}}  & \multicolumn{5}{c}{\textbf{MAG-Scholar-C}}  \\
&Mean $\ell$& $\gamma$ & Time & Memory & Accuracy & Mean $\ell$ & $\gamma$ & Time & Memory & Accuracy\\
\hline
\textbf{PPRGo  } & 32 & -- & 8.8(0.1) & 5.3(0.0)     & 25.88(1.47) & 32 & -- & 254.3(9.8) & \textbf{28.0(0.6)} & \textbf{61.39(3.08)} \\
\textbf{CorePPR  T\&T }& 32 & 0.26(0.10) & \textbf{4.0(11.0)} & 5.8(0.2) & \textbf{33.53(1.07)} &-- &-- &-- &-- &-- \\
\textbf{CorePPR  OT } & 32 & 0.26(0.10) & 5.8(0.0) & 13.3(0.5) & 28.97(1.39) & 32 & 0.45(0.01) & \textbf{238.2(2.8)} & 28.3(0.6) &  60.06(2.89) \\
\textbf{CorePPR  T\&T d${\boldsymbol \ell}$ } & 2  & 0.25(0.11) & 6.0(0.3) & \textbf{5.2(17.2)} & \textbf{33.53(1.01)}  &-- &-- &-- &-- &-- \\
\textbf{CorePPR  OT d${\boldsymbol \ell}$ } & 2 & 0.25(0.11)   & 5.6(0.0)         & 12.7(4.3) & 28.89(1.20)                         & 2               & 0.45(0.02)   &  248.1(10.5) & 28.7(0.6) & 58.02(3.59) \\         \hline
\hline
\end{tabular}}\\
\end{center}
\end{table*}

\end{document}